\documentclass[10pt,twocolumn,letterpaper]{article}

\usepackage{iccv}
\usepackage{times}
\usepackage{epsfig}
\usepackage{graphicx}
\usepackage{amsmath}
\usepackage{amssymb}
\usepackage{bbm}
\usepackage{booktabs}
\usepackage{subcaption}
\usepackage{bm}
\usepackage{multirow}
\usepackage{makecell}
\usepackage{amsfonts}
\usepackage[accsupp]{axessibility}

\newcommand{\stdv}[1]{\scriptsize$\pm$#1}

\usepackage[pagebackref,breaklinks,colorlinks]{hyperref}

\iccvfinalcopy 

\def\iccvPaperID{****} 

\ificcvfinal\fi

\begin{document}

\title{Unilaterally Aggregated Contrastive Learning with Hierarchical Augmentation for Anomaly Detection}

\author{Guodong Wang$^{1,2}$, Yunhong Wang$^2$, Jie Qin$^3$, Dongming Zhang$^4$, Xiuguo Bao$^4$, Di Huang$^{1,2}$\thanks{Corresponding author.}\\
$^1$State Key Laboratory of Software Development Environment, Beihang University, Beijing, China\\
$^2$School of Computer Science and Engineering, Beihang University, Beijing, China\\
$^3$College of Computer Science and Technology, NUAA, Nanjing, China \\
$^4$Natl. Comp. Net. Emer. Resp. Tech. Team/Coord. Ctr. of China, Beijing, China \\
{\tt\small \{wanggd,yhwang,dhuang\}@buaa.edu.cn, qinjiebuaa@gmail.com, zhdm@cert.org.cn, baoxiuguo@139.com
}
}

\maketitle
\ificcvfinal\thispagestyle{empty}\fi


\begin{abstract}
Anomaly detection (AD), aiming to find samples that deviate from the training distribution, is essential in safety-critical applications. Though recent self-supervised learning based attempts achieve promising results by creating virtual outliers, their training objectives are less faithful to AD which requires a concentrated inlier distribution as well as a dispersive outlier distribution. In this paper, we propose \textbf{Uni}laterally Aggregated \textbf{Con}trastive Learning with \textbf{H}ierarchical \textbf{A}ugmentation (\textbf{UniCon-HA}), taking into account both the requirements above. Specifically, we explicitly encourage the concentration of inliers and the dispersion of virtual outliers via supervised and unsupervised contrastive losses, respectively. Considering that standard contrastive data augmentation for generating positive views may induce outliers, we additionally introduce a soft mechanism to re-weight each augmented inlier according to its deviation from the inlier distribution, to ensure a purified concentration. Moreover, to prompt a higher concentration, inspired by curriculum learning, we adopt an easy-to-hard hierarchical augmentation strategy and perform contrastive aggregation at different depths of the network based on the strengths of data augmentation. Our method is evaluated under three AD settings including unlabeled one-class, unlabeled multi-class, and labeled multi-class, demonstrating its consistent superiority over other competitors.
\end{abstract}

\section{Introduction}
\label{sec:intro}

\begin{figure}[!h]
	\centering
	\includegraphics[width=0.5\textwidth]{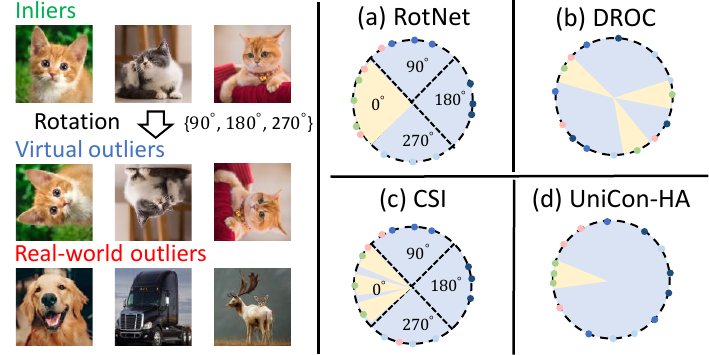}
	\caption{The decision regions (light yellow) of RotNet \cite{hendrycks2019using_self}, DROC~\cite{sohn2020learning}, CSI~\cite{tack2020csi} and our UniCon-HA with rotation used to create virtual outliers. (a) RotNet models the inlier distribution by predicting rotation angles through a 4-way classifier; (b) DROC performs instance discrimination within the union of inliers and their rotations, resulting in a uniform distribution of data points; (c) CSI combines contrastive learning with a rotation classifier, enclosing a sub-region for inliers; (d) Our UniCon-HA explicitly promotes the concentration of inliers and the dispersion of rotated virtual outliers, yielding the most compact decision region.}
	\label{fig:tease}
\end{figure}

Anomaly detection (AD), \emph{a.k.a.} outlier detection, aims to find anomalous observations that considerably deviate from the normality, with a broad range of applications, \emph{e.g.} defect detection \cite{bergmann2019mvtec} and medical diagnosis \cite{schlegl2017unsupervised}. Due to the inaccessibility of real-world outliers, it is typically required to develop outlier detectors solely based on in-distribution data (inliers). Conventional methods \cite{breunig2000lof, latecki2007outlier, kirichenko2020normalizing, ren2019likelihood, zong2018deep, zhai2016deep} leverage generative models to fit the distribution by assigning high densities to inliers; however, they make use of raw images and are fragile caused by background statistics \cite{ren2019likelihood} or pixel correlations \cite{kirichenko2020normalizing}, unexpectedly assigning higher likelihoods to unseen outliers than inliers.

Alternatively, discriminative models \cite{tax2004support, scholkopf1999support, ghafoori2020deep, ruff2018deep} describe the support of the training distribution using classifiers, circumventing the complicated process of density estimation. Furthermore, some studies \cite{golan2018deep, hendrycks2019using_self, bergman2020classification} observe improved performance with the introduction of virtual outliers\footnote{We note that several studies \cite{liang2018enhancing, ruff2019deep, qiu2022latent,wang2022hierarchical} leverage real-world outliers to address AD in a relaxed setting, which is out of the scope of this paper.}, followed by a series of methods \cite{sohn2020learning, tack2020csi, winkens2020contrastive} exploring outliers in a more effective way. For example, classification-based AD methods \cite{golan2018deep, hendrycks2019using_self, bergman2020classification} rely on transformations to generate virtual outliers for creating pretext tasks, with rotation prediction being the most effective. Recently, DROC \cite{sohn2020learning} models the union of inliers and rotated ones via contrastive learning. CSI \cite{tack2020csi} combines contrastive learning with an auxiliary classification head for predicting rotations, further boosting the performance. Overall, these methods deliver better results than their counterparts without using rotation; unfortunately, their imperfect adaptation to AD leaves much room for improvement.

We remark that a good representation distribution for AD requires: (a) a compact distribution for inliers and (b) a dispersive distribution for (virtual) outliers. From our view, the existing methods \cite{hendrycks2019using_self, sohn2020learning, tack2020csi} demonstrate unsatisfactory performance due to the lack of a comprehensive consideration of both aspects. RotNet \cite{hendrycks2019using_self} learns representations which only guarantee that they are distinguishable between labels, \emph{i.e.} four rotation angles $\{0^\circ, 90^\circ, 180^\circ, 270^\circ\}$, resulting in less compact concentration for inliers and less dispersive distribution for outliers shown in Fig.~\ref{fig:tease}(a). Though DROC \cite{sohn2020learning} explicitly enlarges the instance-level distance via contrastive learning and generates a dispersive distribution for outliers, it inevitably pushes inliers away from each other, failing to meet the requirement of a compact inlier distribution \cite{steinwart2005classification} (Fig.~\ref{fig:tease}(b)). CSI \cite{tack2020csi} extends DROC with a rotation classification head which restricts inliers to a sub-region determined by separating hyperplanes, making inliers lumped to some extent (Fig.~\ref{fig:tease}(c)), but the insufficient degree of concentration by the predictor limits its potential.

We also notice a growing trend~\cite{reiss2023mean, guiconstrained, youunified} in leveraging models pre-trained on large-scale datasets (\emph{e.g.} ImageNet \cite{deng2009imagenet}) for AD. However, strictly speaking, they deviate from the objective of AD that outliers stem from an unknown distribution and similar outliers are unseen in training.

In this work, we focus on a strict setting where AD models are trained from scratch using inliers only. We present a novel method for AD based on contrastive learning, named \textbf{Uni}laterally Aggregated \textbf{Con}trastive Learning with \textbf{H}ierarchical \textbf{A}ugmentation (\textbf{UniCon-HA}), to fulfill the goal of inlier concentration and outlier dispersion (Fig.~\ref{fig:tease}(d)). The term \textit{unilaterally} refers to the aggregation applied to inliers only. For inlier concentration, different from other contrastive learning-based AD alternatives \cite{winkens2020contrastive, sehwag2021ssd, tack2020csi, sohn2020learning} that universally perform instance discrimination within the whole training set regardless of inliers or outliers, we take all inliers as one class and pull them together while push outliers away. For outlier dispersion, we perform instance discrimination within all virtual outliers to disperse them around the latent space unoccupied by inliers. Furthermore, considering that the standard augmentation pipeline for generating multiple positive views probably induces outliers as false positives \cite{wang2021improving, bai2022directional} (\emph{e.g.} random crop at background regions), we propose to aggregate augmented views of inliers with a soft mechanism based on the magnitude of deviation from the inlier distribution, with distant samples assigned with lower weights. Finally, to prompt a higher concentration for inliers, inspired by curriculum learning (CL) \cite{bengio2009curriculum}, we adopt an easy-to-hard hierarchical augmentation and perform aggregation at different network depths based on the strengths of data augmentation. Notably, our formulation is free from any auxiliary branches for transformation prediction \cite{tack2020csi, golan2018deep} or pre-trained models \cite{bergmann2020uninformed, fort2021exploring}. We evaluate our method in three typical AD scenarios including one-class, unlabeled multi-class, and labeled multi-class settings. Additionally, the results can be further improved with the introduction of outlier exposure (OE) \cite{hendrycks2019oe}, which is previously deemed harmful in contrastive learning-based CSI \cite{tack2020csi}.

Our main contributions are three-fold:

\begin{itemize}

    \item We present a novel contrastive learning method for AD, simultaneously encouraging the concentration for inliers and the dispersion for outliers, with soft aggregation to suppress the influence of potential outliers induced by data augmentation.

    \item For a higher concentration of inliers, we propose an easy-to-hard hierarchical augmentation strategy and perform contrastive aggregation distributed in the network where deeper layers are responsible for aggregation under stronger augmentations.
    
    \item Experimental results demonstrate the consistent improvement of our method over state-of-the-art competitors under various AD scenarios.
    
\end{itemize}


\section{Related Work}

\textbf{Anomaly Detection.} Recent efforts on AD can be broadly categorized as (a) reconstruction-based \cite{wang2021student, gong2019memorizing, venkataramanan2020attention}, (b) generative \cite{breunig2000lof, latecki2007outlier, ren2019likelihood, kirichenko2020normalizing}, (c) discriminative \cite{tax2004support, scholkopf2001estimating} and (d) self-supervised methods \cite{tack2020csi, hendrycks2019using_self, sohn2020learning}. Generative methods model the density of training data, and examples situated in low-density regions are deemed as outliers. Unfortunately, the curse of dimensionality hinders accurate distribution estimation. Deep generative methods \cite{kirichenko2020normalizing, ren2019likelihood} prove effective in high-dimensional data; however, they work on raw images and still suffer from background statistics \cite{ren2019likelihood} or pixel correlations \cite{kirichenko2020normalizing}.  One-class support vector machine (OC-SVM) \cite{scholkopf2001estimating} and support vector data description (SVDD) \cite{tax2004support} are classic discriminative representatives for AD. While they are powerful with non-linear kernels, their performance is limited to the quality of underlying data representations. Early attempts on AD \cite{tax2004support, scholkopf2001estimating} rely on kernel tricks and hand-crafted feature engineering, but recent ones \cite{ghafoori2020deep, wu2019deep, reiss2021panda, ruff2019deep, ruff2018deep} advocate the capability of deep neural networks to automatically learn high-level representations, outperforming their kernel-based counterparts. However, naive training results in a trivial solution with a constant mapping, \emph{a.k.a.} hypersphere collapse. Previous methods regularize learning by introducing architectural constraints \cite{ruff2018deep}, auto-encoder pre-training \cite{ruff2018deep, ruff2019deep} \emph{etc}, among which introducing outliers \cite{tack2020csi, sohn2020learning, hendrycks2019using_self, golan2018deep, hendrycks2019oe, ruff2019deep, li2021cutpaste} is the most effective to prevent from hypersphere collapse \cite{ruff2018deep}. Building upon the success of self-supervised learning, recent progress \cite{tack2020csi, sohn2020learning, winkens2020contrastive, sehwag2021ssd} is made by adapting contrastive learning to AD with improved performance reported. For example, DROC~\cite{sohn2020learning} and CSI \cite{tack2020csi} leverage distributional augmentation (\emph{e.g.}, rotation) to simulate real-world outliers and model the inlier distribution by contrasting original samples with these simulated outliers. However, the learned representation is uniformly distributed on the hypersphere, contradicting the core principle of AD, which emphasizes that the inlier distribution should remain compact against outliers \cite{ruff2018deep, tax2004support}. Hence, to align contrastive learning more harmoniously with AD, we modify the optimization objective: unlike prior work \cite{tack2020csi, sohn2020learning} performing instance discrimination within all training data (comprising both inliers and virtual outliers), our method explicitly encourages the concentration of inliers and the dispersion of outliers. This adaptation better adheres to the principles of AD.

\textbf{Self-supervised Learning.} Self-supervised learning (SSL), a generic learning framework that automatically generates data labels via either creating pretext tasks or performing contrastive learning, has achieved notable successes in enhancing visual representation learning. Common pretext tasks include predicting image rotations \cite{gidaris2018unsupervised} or patch positions \cite{doersch2015unsupervised}, coloring images \cite{zhang2016colorful} and solving jigsaw puzzles \cite{noroozi2016unsupervised, wang2022jigsaw}, \emph{etc}. In addition to hand-crafted designs for pretext tasks, contrastive learning \cite{he2020momentum, chen2020simple} serves as an alternative in the form of instance discrimination for generic representation learning, benefiting a diversity of downstream vision tasks, such as image recognition and object detection. As opposed to vanilla contrastive learning where each instance itself forms a category, SupCLR \cite{khosla2020supervised}, a supervised extension of contrastive learning, considers multiple positive samples tagged by discriminative labels to help with pulling together intra-class points while pushing apart inter-class ones. This consistently surpasses the performance of the cross-entropy (CE) loss. Aligning with the core concept of AD that inliers are concentrated and outliers are dispersed, this work capitalizes on the advantages of both supervised and unsupervised contrastive learning by explicitly pulling together inliers and pushing apart outliers, respectively.

\begin{figure*}[!h]
	\centering
	\includegraphics[width=1.00\textwidth]{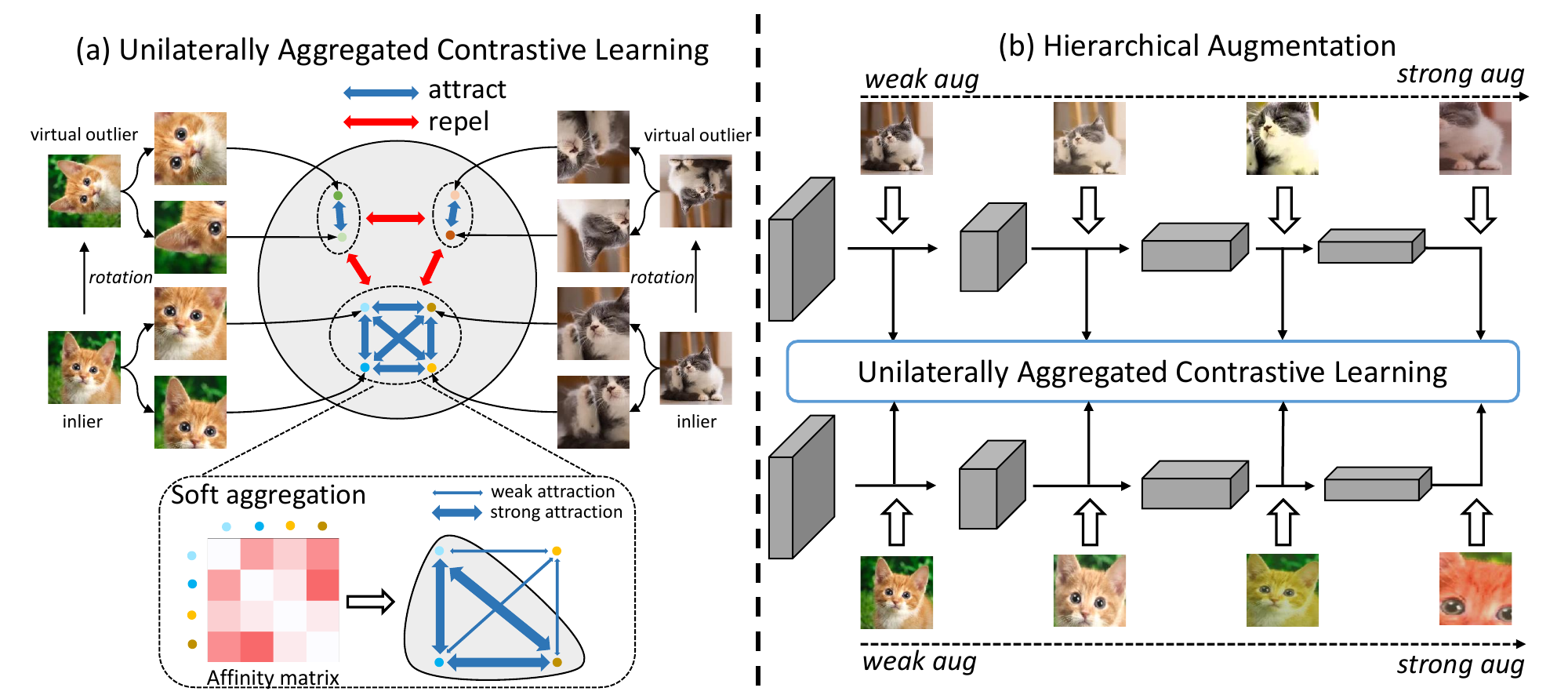}
	\caption{Overview of the proposed UniCon-HA for anomaly detection: (a) we explicitly encourage concentration of inliers and dispersion of virtual outliers generated by distributionally-shifted transformations with rotation being an example. To ensure a purified inlier concentration, we propose soft aggregation to re-weight each view of inliers generated by standard data augmentation \emph{e.g.} random crop, based on its average similarities with all other inliers; (b) to prompt a higher concentration, we employ an easy-to-hard hierarchical augmentation strategy and distribute contrastive aggregation at different stages of the network based on the strengths of data augmentations.}
	\label{fig:overall_arch}
\end{figure*}

\section{Method}

In this section, we first revisit the preliminaries of unsupervised and supervised contrastive learning. Then we introduce our AD method based on contrastive learning, which is specialized to optimize the concentration of inliers and dispersion of virtual outliers along with a soft mechanism to ensure a purified concentration. Moreover, we leverage an easy-to-hard hierarchical augmentation to prompt a higher concentration of inliers along the network layers.

\subsection{Preliminaries}

\textbf{Unsupervised Contrastive Learning.} Unsupervised contrastive learning aims to learn representations from unlabeled data. The premise is that similar samples as positive pairs are supposed to have similar representations. The practical way to create positive pairs is to apply random augmentation to the same sample independently, \emph{e.g.} two crops of the same image or multi-modal views of the same scene. Formally, let $x$ be an anchor, $D_x^+$ and $D_x^-$ be the sets of positive and negative samples w.r.t. $x$, respectively. We consider the following common form of the contrastive loss:
\begin{equation}
\begin{split}
    & {\cal L}_{\rm cons}(x, D_x^+, D_x^-) := \\ 
    & -\frac{1}{|D_x^+|}  \sum_{x^{\prime} \in D_x^+} {   {\rm log} \frac{e^{z(x^{\prime})^T z(x) /\tau}}
      {\sum_{x^{\prime} \in D_x^+ \cup D_x^-  }{e^{z(x^{\prime})^T z(x) /\tau}}    }   },
    \label{eq:cl}
\end{split}
\end{equation}
where $|D_x^+|$ denotes the cardinality of $D_x^+$, $z(\cdot)$ extracts the $\ell_2$-normalized representation of $x$ and $\tau > 0$ is a temperature hyper-parameter. We, in this work, specifically consider the simple contrastive learning framework \emph{i.e.} SimCLR based on instance discrimination. For a batch of unlabeled images ${\cal B} := \{x\}_{i=1}^N$, we first apply a composition of pre-defined identity-preserving augmentations $\cal T$ to construct two views ${\tilde x_i^1}:=t_1(x_i)$ and ${\tilde x_i^2}:=t_2(x_i)$ of the same instance $x_i$, where $t_1, t_2 \sim {\cal T}$. The contrastive loss of SimCLR is defined as follows:

\vspace{-6mm}
\begin{equation}
\begin{split}
    \label{eq:simclr}
    {\cal L}_{\rm SimCLR}({\cal B}; {\cal T})=\frac{1}{2N} \sum_{i=1}^{N}   {\cal L}_{\rm cons}(\tilde x^1_i, \{ \tilde x^2_i \}, {\cal \tilde B} -  \{\tilde x^1_i, \tilde x^2_i\} ) \\  + {\cal L}_{\rm cons}(\tilde x^2_i, \{ \tilde x^1_i \}, {\cal \tilde B} -  \{\tilde x^2_i, \tilde x^1_i\} ), 
\end{split} 
\end{equation}
where ${\cal \tilde B} := {\cal \tilde B}^1 \cup {\cal \tilde B}^2$, ${\cal \tilde B}^1 := \{\tilde x^1\}_{i=1}^N$ and ${\cal \tilde B}^2 := \{\tilde x^2\}_{i=1}^N$. In this case, $D_{\tilde x_i^1}^+ := \{\tilde x_i^2\}$, $D_{\tilde x_i^2}^+ := \{\tilde x_i^1\}$ and $D_{\tilde x_i^2}^- = D_{\tilde x_i^1}^- := {\cal \tilde B} - \{\tilde x_i^2, \tilde x_i^1\}$.

\textbf{Supervised Contrastive Learning.} SupCLR \cite{khosla2020supervised} is a supervised extension of SimCLR by considering class labels. Different from the unsupervised contrastive loss in Eq.~\ref{eq:cl} where each sample has only one positive sample, \emph{i.e.} the augmented view of itself, there are multiple positive samples sharing the same class label, resulting in multiple clusters in the representation space corresponding to their labels. Formally, given a batch of labeled training samples ${\cal C}:=\{(x_i, y_i)\}_{i=1}^N$ with class label $y_i \in {\cal Y}$, ${\cal \tilde C}^1:=\{(\tilde x^1_i, y_i)|{\tilde x}^1_i\in{\cal \tilde B}^1\}_{i=1}^N$ and ${\cal \tilde C}^2:=\{(\tilde x^2_i, y_i)|{\tilde x}^2_i\in{\cal \tilde B}^2 \}_{i=1}^N$ are the two sets of augmented views. The supervised contrastive loss is given as follows:
\begin{equation}
    {\cal L}_{\rm SupCLR}({\cal C}; {\cal T}) = \frac{1}{2N} \sum_{i=1}^{2N} {\cal L}_{\rm cons}({\tilde x_i}, D_{x_i}^+, D_{x_i}^-),
    \label{eq:supclr}
\end{equation}
where $D_{\tilde x_i}^+ := \{x|(x, y_i) \in {\cal \tilde C}^1 \cup {\cal \tilde C}^2   \} - \{ {\tilde x_i}\}$ and $D_{\tilde x_i}^- := {\cal \tilde B} - \{\tilde x_i\} - D_{\tilde x_i}^+$.

\subsection{Unilaterally Aggregated Contrastive Learning} 

Recall that a good representation distribution for AD entails a concentrated grouping of inliers and an appropriate dispersion of outliers. Given that only inliers are available for training, a natural question arises: how to obtain outliers? Due to the inaccessibility of real-world outliers, some attempts are investigated to create virtual outliers, aiming at a trade-off in various manners, such as through transformations \cite{tack2020csi, bergman2020classification, hendrycks2019using_self, li2021cutpaste} or by sourcing them from additional datasets, known as OE \cite{hendrycks2019oe}. These methods, relying on virtual outliers, display the superiority over their counterparts based on inliers only; however, they all fall short in fully addressing both the requirements of a good representation distribution for AD.

Following the success of introducing virtual outliers, in this work, we directly treat the goal of encouraging inlier concentration and outlier dispersion as the optimization objective via a pure contrastive learning framework. We particularly design a novel contrastive loss, namely UniCLR, to unilaterally aggregate inliers and disperse outliers. Different from the existing contrastive learning methods for AD \cite{sohn2020learning, sehwag2021ssd, winkens2020contrastive, tack2020csi} that equally treat each instance from inliers and virtual outliers as one class and perform universal instance discrimination, we take all inliers as one class while each outlier itself a distinct class. 

Formally, given a training inlier set ${\cal D}_{\rm in}$, we first apply distributionally-shifted augmentation ${\cal S}$, \emph{e.g.} rotation, to inliers to create a set of virtual outliers ${\cal D}_{\rm vout} \equiv \{ s(x)|x \in {\cal D}_{\rm in}\ \land s\in{\cal S} \}$. Note that ${\cal D}_{\rm in}$ and ${\cal D}_{\rm vout}$ are disjoint. For each image $x_i \in {\cal D}_{\rm in}/{\cal D}_{\rm vout}$, we further apply identity-preserving augmentations ${\cal T}$ to create two views of $x_i$ and finally obtain ${\cal \tilde D}_{\rm in} := {\cal \tilde D}_{\rm in}^1 \cup  {\cal \tilde D}_{\rm in}^2$ and ${\cal \tilde D}_{\rm vout} := {\cal \tilde D}_{\rm vout}^1 \cup  {\cal \tilde D}_{\rm vout}^2$, based on which we prepare a batch ${\cal \tilde B} := {\cal \tilde D}_{\rm in} \cup {\cal \tilde D}_{\rm vout}$ for training. The contrastive objective is given as:
\begin{equation}
    {\cal L}_{\rm UniCLR}({\cal D}_{\rm in} \cup {\cal D}_{\rm vout}; {\cal T})= \frac{1}{|{\cal \tilde B}|} \sum_{i=1}^{|{\cal  D}_{\rm in}| + |{\cal  D}_{\rm vout}|}{{\cal L}^i_{\rm UniCLR}},
\end{equation}
\begin{equation}
\begin{split}
    & {\cal L}_{\rm UniCLR}^i = \left\{ 
    \begin{aligned}
        & {\cal L}_{\rm cons}({\tilde x_i}^1, {\cal \tilde D}_{\rm in} - \{ {\tilde x_i}^1 \}, {\cal \tilde D}_{\rm vout}) + \\
        & \quad {\cal L}_{\rm cons}({\tilde x_i}^2, {\cal \tilde D}_{\rm in} - \{ {\tilde x_i}^2 \}, {\cal \tilde D}_{\rm vout})
        , x_i \in {\cal  D}_{\rm in}, \\
        & {\cal L}_{\rm cons}({\tilde x_i}^1, \{ {\tilde x_i}^2 \},  {\cal \tilde B} - \{ {\tilde x_i}^2, {\tilde x_i}^1 \} )  + \\
        &  \quad {\cal L}_{\rm cons}({\tilde x_i}^2, \{ {\tilde x_i}^1 \},  {\cal \tilde B} - \{ {\tilde x_i}^1, {\tilde x_i}^2 \} ), x_i \in {\cal D}_{\rm vout}.
    \end{aligned}
    \right.
\end{split}
\label{eq:uniclr}
\end{equation}

Our formulation is structurally similar to SimCLR (Eq.~\ref{eq:simclr}) and SupCLR (Eq.~\ref{eq:supclr}), with some modifications for AD in consideration of both inlier aggregation and outlier dispersion. Though our method is originally designed for inliers without class labels, it can be easily extended to the labeled multi-class setting where inliers sharing the same label are positive views of each other while samples from either other classes or augmented by $\cal S$ are negative.

\textbf{Soft Aggregation.} Data augmentation plays a central role in contrastive representation learning \cite{chen2020simple}. Though the commonly adopted augmentation pipeline in contrastive learning has witnessed the advanced progress in diverse downstream tasks, we observe that excessive distortion applied on the images inevitably shifts the original semantics, inducing outlier-like samples \cite{wang2021improving, bai2022directional}. Aggregating these semantic-drifting samples hinders the inlier concentration. A straightforward solution is to restrict the augmentation strength and apply weak augmentations to inliers; however, it cannot guarantee learning class-separated representations, \emph{i.e.} reliably aggregating different instances with similar semantics \cite{haochen2021provable, wang2021chaos}. To take advantage of the diverse samples by strong augmentations while diminishing the side effects of outliers, we propose to aggregate augmented views of inliers with a soft mechanism based on the magnitude of deviation from the inlier distribution, and follow the notions defined in Eq.~\ref{eq:cl} to formulate it as follows:
\begin{equation}
    \begin{split}
    \label{eq:soft}
    & {\cal L}_{\rm soft\_cons}(x, D_x^+, D_x^-) :=  \\
    & -\frac{1}{  \sum_{x^{\prime} \in D_x^+} {w_x} {w_{x^{\prime}}} } \sum_{x^{\prime}\in D_x^+}{ {\rm log} \frac{{w_x}{w_{x^{\prime}} e^{z(x^{\prime})^T z(x) /\tau}  }}{{\sum_{x^{\prime} \in D_x^+ \cup D_x^-  }{e^{z(x^{\prime})^T z(x) /\tau}} } }},
    \end{split}
\end{equation}
where $w_x$ is the soft weight indicating the importance of sample $x$ in aggregation. According to Eq.~\ref{eq:soft}, a positive pair of $x$ and $x^{\prime}$ receives more attention only if the corresponding $w_x$ and $w_{x^{\prime}}$ are both sufficiently large. Specifically, we measure $w_x$($w_{x^{\prime}}$) for $x(x^\prime)$ by calculating the normalized average similarities with other inliers, \emph{i.e.}  
$D_x^+$, as follows:
\begin{equation}
        \omega_{x_i} = \frac{  \sum_{x_j \in D_x^+ \backslash \{x_i\}} e^{z(x_i)^T z(x_j) /\tau_\omega}  }{    \sum_{x_k \in D_x^+}{\sum_{x_j \in D_x^+ \backslash \{x_k\}}}  e^{z(x_k)^T z(x_j) /\tau_\omega}  },
        \label{eq:w}
\end{equation}
where $\tau_\omega$ controls the sharpness of the weight distribution. Intuitively, if one is away from all other inliers, there is a high probability that it is an outlier and vice versa. We apply soft aggregation (SA) only on inliers, \emph{i.e.}, replacing ${\cal L}_{\rm cons}$ with ${\cal L}_{\rm soft\_cons}$ only for $x_i \in {\cal  D}_{\rm in}$ in Eq.~\ref{eq:uniclr}.

\subsection{Hierarchical Augmentation}

Though the proposed UniCLR is reasonably effective for aggregating inliers and separating them from outliers, one can further improve the performance by prompting a more concentrated distribution for inliers. Inspired by the success of deep supervision in classification, we propose to aggregate inliers with hierarchical augmentation (HA) at different depths of the network based on the strengths of data augmentation. In alignment with the feature extraction process that shallow layers learn low-level features while deep layers emphasize more on high-level task-related semantic features, motivated by curriculum learning (CL) \cite{bengio2009curriculum}, we set stronger augmentation strengths for deeper layers and vice versa, aiming at capturing the distinct representations of inliers from low-level properties and high-level semantics at shallow and deep layers, respectively. To this end, we apply a series of augmentations at different network stages and gradually increase the augmentation strength as the network goes deep. Each stage is responsible for unilaterally aggregating inliers and dispersing virtual outliers generated with the corresponding augmentation strengths.

Formally, we have four sets of augmentation $T_i$, corresponding to four stages $res_i$ in ResNet~\cite{he2016deep}. Each $T_i$ is composed of the same types of augmentations but with different augmentation strengths. Extra projection heads $g_i$ are additionally attached at the end of $res_i$ to down-sample and project the feature maps with the same shape as in the last stage. With $T_1\sim T_4$ applied to inliers ${\cal D}_{\rm in}$ and outliers ${\cal D}_{\rm vout}$, we extract their features $z_i(x)$ with $res_i$ and $g_i$:

\begin{equation}
    z_i(x) = g_i(res_i(T_i(x))), i=1,2,3,4.
\end{equation}

Based on the features extracted by projector $g_i$, we separately perform unilateral aggregation for inliers and dispersion for outliers at each stage. The overall training loss can be formulated as:

\begin{equation}
   {\cal L}_{\mathrm{all}} = \frac{1}{4} \sum_{i=1}^{4}{\lambda_i {\cal L}_{\rm UniCLR} ({\cal D}_{\rm in} \cup {\cal D}_{\rm vout}); T_i), }
\end{equation}
where $\lambda_i$ balances the loss at different network stages. 

Through enforcing supervision in shallow layers, the resulting inlier distribution under the strong augmentation becomes more compact and more distinguishable from outliers.

\begin{table*}
    \vspace{-1mm}
    \begin{subtable}[t]{1.0\linewidth}
        \caption{One-class CIFAR-10.}
        \resizebox{\linewidth}{!}{
      \begin{tabular}{ll|llllllllll|c}
        \toprule
            Method & Network & Plane & Car & Bird & Cat & Deer & Dog & Frog & Horse & Ship & Truck & Mean \\
            \midrule
            AnoGAN \cite{schlegl2017unsupervised} & DCGAN & 67.1 & 54.7 & 52.9 & 54.5 & 65.1 & 60.3 & 58.5 & 62.5 & 75.8 & 66.5 & 61.8 \\
            PLAD \cite{cai2022perturbation} & LeNet & 82.5\stdv{0.4} & 80.8\stdv{0.9} & 68.8\stdv{1.2} & 65.2\stdv{1.2} & 71.6\stdv{1.1} & 71.2\stdv{1.6} & 76.4\stdv{1.9} & 73.5\stdv{1.0} & 80.6\stdv{1.8} & 80.5\stdv{0.3} & 75.1 \\ 
            Geom \cite{golan2018deep} & WRN-16-8 & 74.7 & 95.7 & 78.1 & 72.4 & 87.8 & 87.8 & 83.4 & 95.5 & 93.3 & 91.3 & 86.0 \\
            Rot$^*$ \cite{hendrycks2019using_self} & ResNet-18 & 
            78.3\stdv{0.2} & 94.3\stdv{0.3} & 86.2\stdv{0.4} & 80.8\stdv{0.6} & 89.4\stdv{0.5} & 89.0\stdv{0.4} & 88.9\stdv{0.4} & 95.1\stdv{0.2} & 92.3\stdv{0.3} & 89.7\stdv{0.3} & 88.4 \\
            Rot+Trans$^*$ \cite{hendrycks2019using_self} & ResNet-18 & 
            80.4\stdv{0.3} & 96.4\stdv{0.2} & 85.9\stdv{0.3} & 81.1\stdv{0.5} & 91.3\stdv{0.3} & 89.6\stdv{0.3} & 89.9\stdv{0.3} & 95.9\stdv{0.1} & 95.0\stdv{0.1} & 92.6\stdv{0.2} & 89.8 \\
            GOAD$^*$\cite{bergman2020classification} & ResNet-18 & 75.5\stdv{0.3} & 94.1\stdv{0.3} & 81.8\stdv{0.5} & 72.0\stdv{0.3} & 83.7\stdv{0.9} & 84.4\stdv{0.3} & 82.9\stdv{0.8} & 93.9\stdv{0.3} & 92.9\stdv{0.3} & 89.5\stdv{0.2} & 85.1 \\
            CSI \cite{tack2020csi} & ResNet-18 & 89.9\stdv{0.1} & 99.1\stdv{0.0} & 93.1\stdv{0.2} & 86.4\stdv{0.2} & 93.9\stdv{0.1} & 93.2\stdv{0.2} & 95.1\stdv{0.1} & 98.7\stdv{0.0} & 97.9\stdv{0.0} & 95.5\stdv{0.1} & 94.3 \\
            iDECODe \cite{kaur2022idecode} & WRN-16-8 & 86.5\stdv{0.0} & 98.1\stdv{0.0} & 86.0\stdv{0.5} & 82.6\stdv{0.1} & 90.9\stdv{0.1} &89.2\stdv{0.1} & 88.2\stdv{0.4} & 97.8\stdv{0.1} &97.2\stdv{0.0}&95.5\stdv{0.1} & 91.2 \\
            SSD \cite{sehwag2021ssd} & ResNet-50 & 82.7 & 98.5 & 84.2 & 84.5 & 84.8 & 90.9 & 91.7 & 95.2 & 92.9 & 94.4 & 90.0 \\
            NDA~\cite{chennovelty} & DCGAN & \textbf{98.5} & 76.5 & 79.6 & 79.1 & 92.4 & 71.7 & 97.5 & 69.1 & 98.5 & 75.2 & 84.3 \\            
            \midrule
            UniCon-HA & ResNet-18 & 91.7\stdv{0.1} & 99.2\stdv{0} & 93.9\stdv{0.1} & 89.5\stdv{0.2} & 95.1\stdv{0.1} & 94.1\stdv{0.2} & 96.6\stdv{0.1} & 98.9\stdv{0.0} & 98.1\stdv{0.0} & 96.6\stdv{0.1} & 95.4 \\
            UniCon-HA + OE & ResNet-18 & 94.6\stdv{0.1} & \textbf{99.3}\stdv{0.0} & \textbf{96.2}\stdv{0.1} & \textbf{92.6}\stdv{0.3} & \textbf{96.2}\stdv{0.2} & \textbf{96.6}\stdv{0.1} & \textbf{97.9}\stdv{0.0} & \textbf{99.1}\stdv{0.1} & \textbf{99.0}\stdv{0.0} & \textbf{97.5}\stdv{0.2} & \textbf{96.9} \\
        \bottomrule
        \end{tabular}}
    \end{subtable}	
    \begin{subtable}[t]{0.495\textwidth}
        \centering
		\vspace{0.1in}
        \caption{One-class CIFAR-100 (20 super-classes).}
        \begin{tabular}{ll|c}
        \toprule
            Method & Network & AUROC \\
            \midrule
            GEOM \cite{golan2018deep} & WRN-16-8 & 78.7 \\
            Rot \cite{hendrycks2019using_self} & ResNet-18 & 79.7 \\
            Rot+Trans \cite{hendrycks2019using_self} & ResNet-18 & 79.8 \\
            GOAD \cite{bergman2020classification} & ResNet-18 & 74.5 \\
            CSI \cite{tack2020csi} & ResNet-18 & 89.6 \\ \midrule
            UniCon-HA  & ResNet-18 & \textbf{92.4}  \\  
        \bottomrule
        \end{tabular}    
    \end{subtable}
    \begin{subtable}[t]{0.495\textwidth}
    \centering
	\vspace{0.1in}
    \caption{One-class ImageNet-30.}
     \begin{tabular}{ll|c}
        \toprule
            Method & Network & AUROC \\
            \midrule
            Rot \cite{hendrycks2019using_self} & ResNet-18 & 65.3 \\
            Rot+Attn \cite{hendrycks2019using_self} & ResNet-18 & 81.6 \\
            Rot+Trans+Attn \cite{hendrycks2019using_self} & ResNet-18 & 84.8 \\
            Rot+Trans+Attn+Resize \cite{hendrycks2019using_self} & ResNet-18 & 85.7 \\
            CSI \cite{tack2020csi} & ResNet-18 & 91.6 \\ \midrule
            UniCon-HA  & ResNet-18 & \textbf{93.2}  \\
        \bottomrule
    \end{tabular} 
    \end{subtable}

    \caption{AUROC scores on one-class (a) CIFAR-10, (b) CIFAR-100 (20 super-classes) and (c) ImageNet-30. For CIFAR-10, we report the means and standard deviations of AUROC averaged over five trials. $^*$ denotes the values from CSI~\cite{tack2020csi}. }
    \label{tab:one_class}
    \vspace{-3mm}
\end{table*}

\subsection{Inference}

During testing, we remove all four projection heads $g_i$. While the existing methods \cite{tack2020csi, sehwag2021ssd} depend on specially designed detection score functions to obtain decent results, we observe that using the simple cosine similarity with the nearest one in the learned feature space is sufficiently effective. The detection score $s_i$ for a test example $x_i$ is given as:

\begin{equation}
   s_i(x_i; \{x_m\}) = {\rm \mathop{max}}_{m} \  {\rm cosine}(f(x_i), f(x_m)) ,
\end{equation}
where $\{x_m\}$ denotes the set of training samples and $f(\cdot)$ extracts the $\ell_2$ normalized representation at the end of $res_4$. Following \cite{tack2020csi, bergman2020classification}, we observe improved performance using an ensemble of representations by test-time augmentation.

\section{Experiments}

We compare our method with the state-of-the-art across three AD settings: unlabeled one-class, unlabeled multi-class, and labeled multi-class. Our method is also evaluated on the realistic MvTec-AD dataset~\cite{bergmann2019mvtec}. The area under the receiver operating characteristic curve (AUROC) is adopted as the evaluation metric. 

\subsection{Implementation Details}

We use ResNet-18~\cite{he2016deep} for all experiments to ensure fair comparison with \cite{tack2020csi, sohn2020learning}. Our models are trained from scratch using SGD for 2,048 epochs, and the learning rate is set to 0.01 with a single cycle of cosine learning rate decay. Following \cite{chen2020simple}, we employ a combination of random resized crop, color jittering, horizontal flip and gray-scale with increasing augmentation strengths for $T_1\sim T_4$ to generate positive views while use rotation $\{90^\circ, 180^\circ, 270^\circ \}$ as the default $\cal S$ to create virtual outliers. Thus, ${\cal D}_{\rm in}$ comprises all original training samples and $\lvert{\cal D}_{\rm out} \rvert$  triples $\lvert {\cal D}_{\rm in} \rvert$ by applying $s \in \cal S$ on $x \in {\cal D}_{\rm in}$. We maintain a 1:3 ratio of inliers to virtual outliers during mini-batch training. Detailed augmentation configurations are available in the \textit{supplementary material}. We exclusively apply SA at the last residual stage, \emph{i.e.} $res_4$ where the strongest augmentations are employed. We set $\tau$ and $\tau_\omega$ as 0.5. For OE~\cite{hendrycks2019oe}, we use 80 Million Tiny Images \cite{torralba200880} as the auxiliary dataset, excluding images from CIFAR-10.

\begin{table*}
    \vspace{-3mm}
    \begin{subtable}[t]{1.0\linewidth}
        \caption{Unlabeled CIFAR-10.}
        \resizebox{\linewidth}{!}{
      \begin{tabular}{ll@{\hspace{6pt}}lllllll}
        \toprule
            Method & Network & SVHN & LSUN & ImageNet & LSUN (FIX) & ImageNet (FIX) & CIFAR-100 & Interp. \\
            \midrule
            Rot \cite{hendrycks2019using_self} & ResNet-18 & 97.6\stdv{0.2} & 89.2\stdv{0.7} & 90.5\stdv{0.3} & 77.7\stdv{0.3} & 83.2\stdv{0.1} & 79.0\stdv{0.1} & 64.0\stdv{0.3} \\
            Rot+Trans \cite{hendrycks2019using_self} & ResNet-18 & 97.8\stdv{0.2} & 92.8\stdv{0.9} & 94.2\stdv{0.7} & 81.6\stdv{0.4} & 86.7\stdv{0.1} & 82.3\stdv{0.2} & 68.1\stdv{0.8} \\
            GOAD \cite{bergman2020classification} & ResNet-18 & 96.3\stdv{0.2} & 89.3\stdv{1.5} & 91.8\stdv{1.2} & 78.8\stdv{0.3} & 83.3\stdv{0.1} & 77.2\stdv{0.3} & 59.4\stdv{1.1} \\
            CSI \cite{tack2020csi}  & ResNet-18 & \textbf{99.8}\stdv{0.0} & 97.5\stdv{0.3} & 97.6\stdv{0.3} & 90.3\stdv{0.3} & 93.3\stdv{0.1} & 89.2\stdv{0.1} & 79.3\stdv{0.2} \\
    
            \midrule
            UniCon-HA & ResNet-18 & 99.5\stdv{0.1} & \textbf{98.5}\stdv{0.2} & \textbf{98.3}\stdv{0.2} & 93.3\stdv{0.3} & 97.8\stdv{0.1} & 90.3\stdv{0.3}  & \textbf{80.7}\stdv{0.2} \\
    
            UniCon-HA + OE & ResNet-18 & 99.2\stdv{0.0} & 97.8\stdv{0.2} & 95.8\stdv{0.1} & \textbf{95.8}\stdv{0.4} & \textbf{98.3}\stdv{0.2} & \textbf{91.6}\stdv{0.2}  & 80.1\stdv{0.1} \\
            \bottomrule
         
        \end{tabular}
        }
    \end{subtable}	
    
    \begin{subtable}[t]{1.0\textwidth}
        \centering
        \caption{Unlabeled ImageNet-30.}
        \begin{tabular}{ll@{\hspace{6pt}}llllllll}
        \toprule
            Method & Network & CUB-200 & Dogs & Pets & Flowers & Food-101 & Places-365 & Caltech-256 & DTD \\
            \midrule
            Rot \cite{hendrycks2019using_self} & ResNet-18 & 76.5 & 77.2 & 70.0 & 87.2 & 72.7 & 52.6 & 70.9 & 89.9 \\
            Rot+Trans \cite{hendrycks2019using_self} & ResNet-18 & 74.5 & 77.8 & 70.0 & 86.3  & 71.6 & 53.1 & 70.0 & 89.4  \\
            GOAD \cite{bergman2020classification} & ResNet-18 &  71.5 & 74.3 & 65.5 & 82.8 & 68.7 & 51.0 & 67.4 & 87.5  \\
            CSI \cite{tack2020csi} & ResNet-18 &  90.5 & 97.1 & 85.2 &94.7 &89.2 & 78.3 & 87.1 & \textbf{96.9} \\
    
            \midrule
            UniCon-HA  & ResNet-18 & \textbf{91.2} & \textbf{97.4} & \textbf{88.0} &  \textbf{95.1} &  \textbf{91.2} & \textbf{84.5} & \textbf{89.6} & 96.5 \\
            \bottomrule
        \end{tabular}
    \end{subtable}
    \caption{AUROC scores on unlabeled (a) CIFAR-10 and (b) ImageNet-30. For CIFAR-10, we report the means and standard deviations of AUROC averaged over five trials.}
    \vspace{-4mm}
    \label{tab:unlabeled_multiclass}
\end{table*}

\subsection{Results}

\textbf{Unlabeled One-class.} In this setting, a single class serves as the inlier, while the remaining classes act as outliers. Following \cite{hendrycks2019using_self, tack2020csi, sohn2020learning, bergman2020classification}, the experiments are performed on CIFAR-10~\cite{krizhevsky2009learning}, CIFAR-100 (20 super-classes)~\cite{krizhevsky2009learning} and ImageNet-30~\cite{hendrycks2019using_self}. In Tab.~\ref{tab:one_class}, we present a comprehensive comparison of our method with a range of alternatives including one-class classifiers, reconstruction-based methods and SSL methods. Notably, SSL methods using virtual outliers generated by shifting transformations such as rotation and translation, yield favorable results compared to those specifically tailored for one-class learning. Thanks to UniCLR with HA, we achieve enhanced performance across all the three datasets. Moreover, introducing supervision through Outlier Exposure (OE) \cite{hendrycks2019oe} nearly solves the CIFAR-10 task, which is previously regarded as less effective in the contrastive based AD method \cite{tack2020csi}. We attribute the success to our contrastive aggregation strategy, which shapes a more focused inlier distribution when more outliers introduced.

\textbf{Unlabeled Multi-class.} This setting expands the one-class dataset to a multi-class scenario, wherein images from different datasets are treated as outliers. In the case of CIFAR-10 as the inlier dataset, we consider SVHN \cite{netzer2011reading}, CIFAR-100~\cite{krizhevsky2009learning}, ImageNet~\cite{liang2018enhancing}, LSUN~\cite{yu2015lsun}, ImageNet (Fix), LSUN (Fix) and linearly-interpolated samples of CIFAR-10 (Interp.) \cite{du2019implicit} as potential outliers. ImageNet (Fix) and LSUN (Fix) are the modified versions of ImageNet and LSUN, designed to address easily detectable artifacts resulting from resizing operations. For ImageNet-30, we consider CUB-200~\cite{wah2011caltech}, Dogs~\cite{khosla2011novel}, Pets~\cite{parkhi2012cats}, Flowers~\cite{nilsback2006visual}, Food-101~\cite{bossard2014food}, Places-365~\cite{zhou2017places}, Caltech256~\cite{griffin2007caltech} and DTD~\cite{cimpoi2014describing} as outlier. Tab.~\ref{tab:unlabeled_multiclass} shows that our UniCon-HA outperforms other counterparts on most benchmarks. Though the training set follows a multi-center distribution, the straightforward aggregation of all data into a single center proves remarkably effective in AD.

\textbf{Labeled Multi-class.} In the multi-class setting with labeled data, rather than treating all inliers as a single class, as seen in the previous scenarios, we designate inliers sharing identical labels as positives. Conversely, inliers with differing labels or those generated by distributionally-shifted augmentations are negatives. From Tab.~\ref{tab:labeled_multiclass}, by incorporating labels into the UniCLR loss, our method not only improves the performance in unlabeled multi-class setting but also consistently surpasses other competitors that employ virtual outliers, \emph{i.e.} RotNet \cite{hendrycks2019using_self}, GEOM \cite{golan2018deep}, CSI \cite{tack2020csi} and DROC \cite{sohn2020learning}. It suggests that our method generalizes well to labeled multi-class inliers.

\textbf{Realistic Dataset.} Following DROC~\cite{sohn2020learning}, we learn patch representations of 32$\times$32. Tab.~\ref{tab:mvtec} shows that our method outperforms the counterparts that also incorporate rotation augmentation. Though CutPaste~\cite{li2021cutpaste} exhibits better performance than ours, it is crucial to understand that CutPaste is specially designed for industrial anomaly localization, making it unsuitable for our settings. For instance, CutPaste only achieves 69.4\% while ours reaches 95.4\% in the one-class CIFAR-10 scenario.

\begin{table}[htp]
    \vspace{-1mm}
    \centering
    \resizebox{0.99\linewidth}{!}{
        \begin{tabular}{ccccccc}
        \toprule
         Level & RotNet~\cite{hendrycks2019using_self} & DROC~\cite{sohn2020learning} & CutPaste~\cite{li2021cutpaste} & \textbf{Ours}  \\ \hline 
         Image    &  71.0 & 86.5 & \textbf{95.2} & 89.8 \\
         Pixel    &  92.6 & 90.4 & \textbf{96.0} & 94.3 \\
         \bottomrule
        \end{tabular}
    }
    \caption{Image/pixel-level AUROC scores on MVTec-AD.}
    \label{tab:mvtec}
    \vspace{-5mm}
\end{table}

\begin{table*}
    \vspace{-3mm}
    \begin{subtable}[t]{1.0\linewidth}
        \caption{Labeled CIFAR-10.}
        \resizebox{\linewidth}{!}{
         \begin{tabular}{lcccccccc}
        \toprule
            Method & Network & SVHN & LSUN & ImageNet & LSUN (FIX) & ImageNet (FIX) & CIFAR100 & Interp. \\
            \midrule
            
            SupCLR \cite{khosla2020supervised}  & ResNet-18 & 97.3\stdv{0.1} & 92.8\stdv{0.5} & 91.4\stdv{1.2} & 91.6\stdv{1.5} & 90.5\stdv{0.5} & 88.6\stdv{0.2} & 75.7\stdv{0.1} \\
            CSI \cite{tack2020csi}  & ResNet-18 & 97.9\stdv{0.1} & 97.7\stdv{0.4} & 97.6\stdv{0.3} & 93.5\stdv{0.4} & 94.0\stdv{0.1} & 92.2\stdv{0.1} & 80.1\stdv{0.3} \\
            \midrule
            
            UniCon-HA  & ResNet-18 & \textbf{99.8}\stdv{0.1} & \textbf{99.1}\stdv{0.2} & \textbf{99.0}\stdv{0.1} & 94.2\stdv{0.3} & 97.9\stdv{0.4} & 92.9\stdv{0.2} & 83.4\stdv{0.3} \\
            UniCon-HA + OE  & ResNet-18  & 98.8\stdv{0.2} & 98.6\stdv{0.3} & 97.9\stdv{0.2} & \textbf{95.5}\stdv{0.3} & \textbf{98.2}\stdv{0.2} & \textbf{93.4}\stdv{0.2} & \textbf{83.5}\stdv{0.1} \\
    
            \bottomrule
        \end{tabular}
        }
    \end{subtable}	

    \vspace{-4mm}
    
    \begin{subtable}[t]{1.0\textwidth}
        \centering
		\vspace{0.1in}
        \caption{Labeled ImageNet-30.}
       \begin{tabular}{ll@{\hspace{6pt}}llllllll}
        \toprule
            Method & Network & CUB-200 & Dogs & Pets & Flowers & Food-101 & Places-365 & Caltech-256 & DTD \\
            \midrule
            Rot \cite{hendrycks2019using_self} & ResNet-18 & 88.0 & 96.7 & 95.0 &  89.7 &  79.8 &  90.5 &  90.6 & 90.1 \\
            Rot+Trans \cite{hendrycks2019using_self} & ResNet-18 &  86.3 & 95.6 & 94.2 & 92.2 & 81.2 & 89.7 & 90.2 & 92.1 \\
            GOAD \cite{bergman2020classification} & ResNet-18 &  93.4 & 97.7 & 96.9 & 96.0 & 87.0 & 92.5 &  91.9 & 93.7  \\
            CSI \cite{tack2020csi} & ResNet-18 &  94.6  &  \textbf{98.3} &  97.4 &  96.2  & 88.9  & 94.0  & 93.2  & 97.4 \\
            \midrule
            UniCon-HA & ResNet-18 & \textbf{94.9} &98.1&\textbf{97.8}&\textbf{96.7}&\textbf{90.9}&\textbf{94.6}&\textbf{95.2}&\textbf{97.7}\\
            \bottomrule
        \end{tabular}
    \end{subtable}
    \caption{AUROC scores on labeled (a) CIFAR-10 and (b) ImageNet-30. For CIFAR-10, we report the means and standard deviations of AUROC averaged over five trials.}
    \label{tab:labeled_multiclass}
    \vspace{-4mm}
\end{table*}

\begin{table}[htp]
    \centering
    \resizebox{0.99\linewidth}{!}{
    \begin{tabular}{cccccc}
    \toprule
      Method & Perm & Sobel & Noise & Blur & Rotation \\  \midrule
      CSI~\cite{tack2020csi}    &  90.7 & 88.3 & 89.3  & 89.2 & 94.3 \\
      UniCon-HA    &  92.1 & 90.4  & 90.8  & 89.8 & \textbf{95.4}  \\
    \bottomrule
    \end{tabular}
    }
    \caption{Ablation study for shifting transformations. Mean AUROC (\%) values are reported on one-class CIFAR-10.}
    \label{tab:oc-cifar10_shifting}
\end{table}

\begin{table}[!ht]
	\centering
    \resizebox{0.99\linewidth}{!}{
    
    	\begin{tabular}{c|c|c|c|c|c|c}
        \toprule
            \multirow{2}{*}{Row} & \multirow{2}{*}{SA} & \multirow{2}{*}{HA} & \multirow{2}{*}{Rot. Cls.} & One-class & \multicolumn{2}{c}{Multi-class CIFAR-10}  \\ \cline{5-7}
            & & & &   CIFAR-10 & Unlabeled & Labeled \\ \hline

            1 & & & \checkmark & 94.3 & 92.4 & 93.3 \\
            2 &  & 4 & \checkmark & 94.6 & 92.8 & 93.9 \\
            3 & & 2-3-4 & \checkmark & 95.1 & 93.7 & 95.0 \\
            4 & \checkmark & 2-3-4 & \checkmark & 95.3 & \textbf{94.2} & \textbf{95.4} \\ \hline

            5 & & & & 92.4 & 89.4 & 90.6 \\
            6 & & 4 & & 94.8 & 93.0 & 93.7 \\
            7 & \checkmark & 4 & & 95.0 & 93.3 & 93.9 \\
            8 & & 2-3-4 & & 95.1 & 93.8 & 94.7 \\
            9 & \checkmark & 2-3-4 & & \textbf{95.4} & 94.1 & 95.2 \\
        \bottomrule
    	\end{tabular}
     
     }
    \caption{Ablation study for SA and HA. Numbers in HA denote the residual stage(s) performing aggregation.}
    \vspace{-2mm}
    \label{tab:hcc}
\end{table}

\subsection{Ablation Study}

We conduct ablation studies on (a) various shifting transformations and (b) aggregation strategies: SA and HA.

\textbf{Shifting Transformation.} In contrast to CSI \cite{tack2020csi}, we investigate various shifting transformations beyond rotation, including CutPerm \cite{tack2020csi}, Gaussian blur, Gaussian noise and Sobel filtering. From Tab.~\ref{tab:oc-cifar10_shifting}, our method consistently outperforms CSI under different shifting transformations, with rotation being the most effective. One plausible explanation is that rotation creates more distinguishable negative samples from the original ones, facilitating the learning process. Notably, different from CSI~\cite{tack2020csi} and RotNet~\cite{hendrycks2019using_self}, we do not learn to differentiate specific transformation types. It encourages the community to rethink the necessity of transformation prediction through a classifier, such as the task of 4-way rotation prediction. Please refer to the \textit{supplementary material} for further analysis on additional transformations.

\textbf{Aggregation Strategy.}
Beyond using the UniCLR loss, we also employ the SA and HA strategies to prompt a more purified and compact concentration of inliers, respectively. To assess the efficacy of each strategy, we establish two baselines: one with a rotation classifier and one without, conducting vanilla contrastive learning on the union of inliers and virtual outliers. The results in Tab.~\ref{tab:hcc} indicate that both single-stage and multi-stage aggregations yield improved outcomes through SA. This underscores the benefits of mitigating the impact of the outliers generated by unexpected data augmentation, thereby purifying the inlier distribution. Tab.~\ref{tab:hcc} reveals two pivotal  observations regarding HA: firstly, the presence of UniCLR diminishes the impact of a rotation classifier (2,3,4 \textit{vs.} 6,8,9), thanks to promoting inlier concentration and outlier dispersion. Secondly, enabling solely $res_4$ for contrastive aggregation significantly improves baselines (1 \textit{vs.} 2, or 5 \textit{vs.} 6). Broadly, HA leads to a noteworthy and consistent gain when applied across more stages (2 \textit{vs.} 3, or 6 \textit{vs.} 8).

\section{Conclusion}
In this work, we address AD with only access to normal images during training. We underline that the concentration of inliers and the dispersion of outliers are two critical factors, which are achieved by a supervised and unsupervised contrastive loss, respectively. To ensure a purified inlier concentration, we propose a soft mechanism to re-weight each view of inliers generated by data augmentation based on its deviation from the inlier distribution. To further prompt a compact inlier concentration, we adopt an easy-to-hard HA and perform aggregation at different network depths based on augmentation strengths. Experiments on three typical AD settings with different benchmarks demonstrate the superiority of our method.

\noindent \textbf{Acknowledgments.} This work is partly supported by the National Key R\&D Program of China (2021ZD0110503), the National Natural Science Foundation of China (62022011 and 62276129),  the Research Program of State Key Laboratory of Software Development Environment, and the Fundamental Research Funds for the Central Universities.


{\small
\bibliographystyle{ieee_fullname}
\bibliography{egbib}

\begin{thebibliography}{10}\itemsep=-1pt

\bibitem{bai2022directional}
Yalong Bai, Yifan Yang, Wei Zhang, and Tao Mei.
\newblock Directional self-supervised learning for heavy image augmentations.
\newblock In {\em CVPR}, 2022.

\bibitem{bengio2009curriculum}
Yoshua Bengio, J{\'e}r{\^o}me Louradour, Ronan Collobert, and Jason Weston.
\newblock Curriculum learning.
\newblock In {\em ICML}, 2009.

\bibitem{bergman2020classification}
Liron Bergman and Yedid Hoshen.
\newblock Classification-based anomaly detection for general data.
\newblock In {\em ICLR}, 2020.

\bibitem{bergmann2019mvtec}
Paul Bergmann, Michael Fauser, David Sattlegger, and Carsten Steger.
\newblock Mvtec ad--a comprehensive real-world dataset for unsupervised anomaly
  detection.
\newblock In {\em CVPR}, 2019.

\bibitem{bergmann2020uninformed}
Paul Bergmann, Michael Fauser, David Sattlegger, and Carsten Steger.
\newblock Uninformed students: Student-teacher anomaly detection with
  discriminative latent embeddings.
\newblock In {\em CVPR}, 2020.

\bibitem{bossard2014food}
Lukas Bossard, Matthieu Guillaumin, and Luc~Van Gool.
\newblock Food-101--mining discriminative components with random forests.
\newblock In {\em ECCV}, 2014.

\bibitem{breunig2000lof}
Markus~M Breunig, Hans-Peter Kriegel, Raymond~T Ng, and J{\"o}rg Sander.
\newblock Lof: identifying density-based local outliers.
\newblock In {\em ACM SIGMOD}, 2000.

\bibitem{cai2022perturbation}
Jinyu Cai and Jicong Fan.
\newblock Perturbation learning based anomaly detection.
\newblock In {\em NeurIPS}, 2022.

\bibitem{chennovelty}
Chengwei Chen, Yuan Xie, Shaohui Lin, Ruizhi Qiao, Jian Zhou, Xin Tan, Yi
  Zhang, and Lizhuang Ma.
\newblock Novelty detection via contrastive learning with negative data
  augmentation.
\newblock In {\em IJCAI}, 2021.

\bibitem{chen2020simple}
Ting Chen, Simon Kornblith, Mohammad Norouzi, and Geoffrey Hinton.
\newblock A simple framework for contrastive learning of visual
  representations.
\newblock In {\em ICML}, 2020.

\bibitem{cimpoi2014describing}
Mircea Cimpoi, Subhransu Maji, Iasonas Kokkinos, Sammy Mohamed, and Andrea
  Vedaldi.
\newblock Describing textures in the wild.
\newblock In {\em CVPR}, 2014.

\bibitem{deng2009imagenet}
Jia Deng, Wei Dong, Richard Socher, Li-Jia Li, Kai Li, and Li Fei-Fei.
\newblock Imagenet: A large-scale hierarchical image database.
\newblock In {\em CVPR}, 2009.

\bibitem{doersch2015unsupervised}
Carl Doersch, Abhinav Gupta, and Alexei~A Efros.
\newblock Unsupervised visual representation learning by context prediction.
\newblock In {\em ICCV}, 2015.

\bibitem{du2019implicit}
Yilun Du and Igor Mordatch.
\newblock Implicit generation and modeling with energy based models.
\newblock In {\em NeurIPS}, 2019.

\bibitem{fort2021exploring}
Stanislav Fort, Jie Ren, and Balaji Lakshminarayanan.
\newblock Exploring the limits of out-of-distribution detection.
\newblock In {\em NeurIPS}, 2021.

\bibitem{ghafoori2020deep}
Zahra Ghafoori and Christopher Leckie.
\newblock Deep multi-sphere support vector data description.
\newblock In {\em SDM}, 2020.

\bibitem{gidaris2018unsupervised}
Spyros Gidaris, Praveer Singh, and Nikos Komodakis.
\newblock Unsupervised representation learning by predicting image rotations.
\newblock In {\em ICLR}, 2018.

\bibitem{golan2018deep}
Izhak Golan and Ran El-Yaniv.
\newblock Deep anomaly detection using geometric transformations.
\newblock In {\em NeurIPS}, 2018.

\bibitem{gong2019memorizing}
Dong Gong, Lingqiao Liu, Vuong Le, Budhaditya Saha, Moussa~Reda Mansour, Svetha
  Venkatesh, and Anton van~den Hengel.
\newblock Memorizing normality to detect anomaly: Memory-augmented deep
  autoencoder for unsupervised anomaly detection.
\newblock In {\em ICCV}, 2019.

\bibitem{griffin2007caltech}
Gregory Griffin, Alex Holub, and Pietro Perona.
\newblock Caltech-256 object category dataset.
\newblock California Institute of Technology, 2007.

\bibitem{guiconstrained}
Xingtai Gui, Yang~Chang Di~Wu, and Shicai Fan.
\newblock Constrained adaptive projection with pretrained features for anomaly
  detection.
\newblock 2022.

\bibitem{haochen2021provable}
Jeff~Z HaoChen, Colin Wei, Adrien Gaidon, and Tengyu Ma.
\newblock Provable guarantees for self-supervised deep learning with spectral
  contrastive loss.
\newblock In {\em NeurIPS}, 2021.

\bibitem{he2020momentum}
Kaiming He, Haoqi Fan, Yuxin Wu, Saining Xie, and Ross Girshick.
\newblock Momentum contrast for unsupervised visual representation learning.
\newblock In {\em CVPR}, 2020.

\bibitem{he2016deep}
Kaiming He, Xiangyu Zhang, Shaoqing Ren, and Jian Sun.
\newblock Deep residual learning for image recognition.
\newblock In {\em CVPR}, 2016.

\bibitem{hendrycks2019oe}
Dan Hendrycks, Mantas Mazeika, and Thomas Dietterich.
\newblock Deep anomaly detection with outlier exposure.
\newblock In {\em ICLR}, 2019.

\bibitem{hendrycks2019using_self}
Dan Hendrycks, Mantas Mazeika, Saurav Kadavath, and Dawn Song.
\newblock Using self-supervised learning can improve model robustness and
  uncertainty.
\newblock In {\em NeurIPS}, 2019.

\bibitem{kaur2022idecode}
Ramneet Kaur, Susmit Jha, Anirban Roy, Sangdon Park, Edgar Dobriban, Oleg
  Sokolsky, and Insup Lee.
\newblock idecode: In-distribution equivariance for conformal
  out-of-distribution detection.
\newblock In {\em AAAI}, 2022.

\bibitem{khosla2011novel}
Aditya Khosla, Nityananda Jayadevaprakash, Bangpeng Yao, and Fei-Fei Li.
\newblock Novel dataset for fine-grained image categorization: Stanford dogs.
\newblock In {\em CVPR workshop}, 2011.

\bibitem{khosla2020supervised}
Prannay Khosla, Piotr Teterwak, Chen Wang, Aaron Sarna, Yonglong Tian, Phillip
  Isola, Aaron Maschinot, Ce Liu, and Dilip Krishnan.
\newblock Supervised contrastive learning.
\newblock In {\em NeurIPS}, 2020.

\bibitem{kirichenko2020normalizing}
Polina Kirichenko, Pavel Izmailov, and Andrew~G Wilson.
\newblock Why normalizing flows fail to detect out-of-distribution data.
\newblock In {\em NeurIPS}, 2020.

\bibitem{krizhevsky2009learning}
Alex Krizhevsky, Geoffrey Hinton, et~al.
\newblock Learning multiple layers of features from tiny images.
\newblock 2009.

\bibitem{latecki2007outlier}
Longin~Jan Latecki, Aleksandar Lazarevic, and Dragoljub Pokrajac.
\newblock Outlier detection with kernel density functions.
\newblock In {\em MLDM}, 2007.

\bibitem{li2021cutpaste}
Chun-Liang Li, Kihyuk Sohn, Jinsung Yoon, and Tomas Pfister.
\newblock Cutpaste: Self-supervised learning for anomaly detection and
  localization.
\newblock In {\em CVPR}, 2021.

\bibitem{liang2018enhancing}
Shiyu Liang, Yixuan Li, and R Srikant.
\newblock Enhancing the reliability of out-of-distribution image detection in
  neural networks.
\newblock In {\em ICLR}, 2018.

\bibitem{netzer2011reading}
Yuval Netzer, Tao Wang, Adam Coates, Alessandro Bissacco, Bo Wu, and Andrew~Y
  Ng.
\newblock Reading digits in natural images with unsupervised feature learning.
\newblock In {\em NeurIPS workshop}, 2011.

\bibitem{nilsback2006visual}
M-E Nilsback and Andrew Zisserman.
\newblock A visual vocabulary for flower classification.
\newblock In {\em CVPR}, 2006.

\bibitem{noroozi2016unsupervised}
Mehdi Noroozi and Paolo Favaro.
\newblock Unsupervised learning of visual representations by solving jigsaw
  puzzles.
\newblock In {\em ECCV}, 2016.

\bibitem{parkhi2012cats}
Omkar~M Parkhi, Andrea Vedaldi, Andrew Zisserman, and CV Jawahar.
\newblock Cats and dogs.
\newblock In {\em CVPR}, 2012.

\bibitem{qiu2022latent}
Chen Qiu, Aodong Li, Marius Kloft, Maja Rudolph, and Stephan Mandt.
\newblock Latent outlier exposure for anomaly detection with contaminated data.
\newblock In {\em ICML}, 2022.

\bibitem{reiss2021panda}
Tal Reiss, Niv Cohen, Liron Bergman, and Yedid Hoshen.
\newblock Panda: Adapting pretrained features for anomaly detection and
  segmentation.
\newblock In {\em CVPR}, 2021.

\bibitem{reiss2023mean}
Tal Reiss and Yedid Hoshen.
\newblock Mean-shifted contrastive loss for anomaly detection.
\newblock In {\em AAAI}, 2023.

\bibitem{ren2019likelihood}
Jie Ren, Peter~J Liu, Emily Fertig, Jasper Snoek, Ryan Poplin, Mark Depristo,
  Joshua Dillon, and Balaji Lakshminarayanan.
\newblock Likelihood ratios for out-of-distribution detection.
\newblock In {\em NeurIPS}, 2019.

\bibitem{ruff2018deep}
Lukas Ruff, Robert Vandermeulen, Nico Goernitz, Lucas Deecke, Shoaib~Ahmed
  Siddiqui, Alexander Binder, Emmanuel M{\"u}ller, and Marius Kloft.
\newblock Deep one-class classification.
\newblock In {\em ICML}, 2018.

\bibitem{ruff2019deep}
Lukas Ruff, Robert~A Vandermeulen, Nico G{\"o}rnitz, Alexander Binder, Emmanuel
  M{\"u}ller, Klaus-Robert M{\"u}ller, and Marius Kloft.
\newblock Deep semi-supervised anomaly detection.
\newblock In {\em ICLR}, 2019.

\bibitem{schlegl2017unsupervised}
Thomas Schlegl, Philipp Seeb{\"o}ck, Sebastian~M Waldstein, Ursula
  Schmidt-Erfurth, and Georg Langs.
\newblock Unsupervised anomaly detection with generative adversarial networks
  to guide marker discovery.
\newblock In {\em IPMI}, 2017.

\bibitem{scholkopf2001estimating}
Bernhard Sch{\"o}lkopf, John~C Platt, John Shawe-Taylor, Alex~J Smola, and
  Robert~C Williamson.
\newblock Estimating the support of a high-dimensional distribution.
\newblock {\em Neural computation}, 13(7):1443--1471, 2001.

\bibitem{scholkopf1999support}
Bernhard Sch{\"o}lkopf, Robert~C Williamson, Alex Smola, John Shawe-Taylor, and
  John Platt.
\newblock Support vector method for novelty detection.
\newblock In {\em NeurIPS}, 2000.

\bibitem{sehwag2021ssd}
Vikash Sehwag, Mung Chiang, and Prateek Mittal.
\newblock Ssd: A unified framework for self-supervised outlier detection.
\newblock {\em ICLR}, 2021.

\bibitem{sohn2020learning}
Kihyuk Sohn, Chun-Liang Li, Jinsung Yoon, Minho Jin, and Tomas Pfister.
\newblock Learning and evaluating representations for deep one-class
  classification.
\newblock In {\em ICLR}, 2020.

\bibitem{steinwart2005classification}
Ingo Steinwart, Don Hush, and Clint Scovel.
\newblock A classification framework for anomaly detection.
\newblock {\em JMLR}, 6(2):211--232, 2005.

\bibitem{tack2020csi}
Jihoon Tack, Sangwoo Mo, Jongheon Jeong, and Jinwoo Shin.
\newblock Csi: Novelty detection via contrastive learning on distributionally
  shifted instances.
\newblock In {\em NeurIPS}, 2020.

\bibitem{tax2004support}
David~MJ Tax and Robert~PW Duin.
\newblock Support vector data description.
\newblock {\em Machine learning}, 54(1):45--66, 2004.

\bibitem{torralba200880}
Antonio Torralba, Rob Fergus, and William~T Freeman.
\newblock 80 million tiny images: A large data set for nonparametric object and
  scene recognition.
\newblock {\em IEEE TPAMI}, 30(11):1958--1970, 2008.

\bibitem{venkataramanan2020attention}
Shashanka Venkataramanan, Kuan-Chuan Peng, Rajat~Vikram Singh, and Abhijit
  Mahalanobis.
\newblock Attention guided anomaly localization in images.
\newblock In {\em ECCV}, 2020.

\bibitem{wah2011caltech}
Catherine Wah, Steve Branson, Peter Welinder, Pietro Perona, and Serge
  Belongie.
\newblock The caltech-ucsd birds-200-2011 dataset.
\newblock California Institute of Technology, 2011.

\bibitem{wang2021student}
Guodong Wang, Shumin Han, Errui Ding, and Di Huang.
\newblock Student-teacher feature pyramid matching for anomaly detection.
\newblock In {\em BMVC}, 2021.

\bibitem{wang2022jigsaw}
Guodong Wang, Yunhong Wang, Jie Qin, Dongming Zhang, Xiuguo Bao, and Di Huang.
\newblock Video anomaly detection by solving decoupled spatio-temporal jigsaw
  puzzles.
\newblock In {\em ECCV}, 2022.

\bibitem{wang2022hierarchical}
Gaoang Wang, Yibing Zhan, Xinchao Wang, Mingli Song, and Klara Nahrstedt.
\newblock Hierarchical semi-supervised contrastive learning for
  contamination-resistant anomaly detection.
\newblock In {\em ECCV}, 2022.

\bibitem{wang2021improving}
Yu Wang, Jingyang Lin, Jingjing Zou, Yingwei Pan, Ting Yao, and Tao Mei.
\newblock Improving self-supervised learning with automated unsupervised
  outlier arbitration.
\newblock In {\em NeurIPS}, 2021.

\bibitem{wang2021chaos}
Yifei Wang, Qi Zhang, Yisen Wang, Jiansheng Yang, and Zhouchen Lin.
\newblock Chaos is a ladder: A new theoretical understanding of contrastive
  learning via augmentation overlap.
\newblock In {\em ICLR}, 2021.

\bibitem{winkens2020contrastive}
Jim Winkens, Rudy Bunel, Abhijit~Guha Roy, Robert Stanforth, Vivek Natarajan,
  Joseph~R Ledsam, Patricia MacWilliams, Pushmeet Kohli, Alan Karthikesalingam,
  Simon Kohl, et~al.
\newblock Contrastive training for improved out-of-distribution detection.
\newblock {\em arXiv preprint arXiv:2007.05566}, 2020.

\bibitem{wu2019deep}
Peng Wu, Jing Liu, and Fang Shen.
\newblock A deep one-class neural network for anomalous event detection in
  complex scenes.
\newblock {\em IEEE TNNLS}, 31(7):2609--2622, 2019.

\bibitem{youunified}
Zhiyuan {You} et~al.
\newblock A unified model for multi-class anomaly detection.
\newblock In {\em NeurIPS}, 2022.

\bibitem{yu2015lsun}
Fisher Yu, Ari Seff, Yinda Zhang, Shuran Song, Thomas Funkhouser, and Jianxiong
  Xiao.
\newblock Lsun: Construction of a large-scale image dataset using deep learning
  with humans in the loop.
\newblock {\em arXiv preprint arXiv:1506.03365}, 2015.

\bibitem{zhai2016deep}
Shuangfei Zhai, Yu Cheng, Weining Lu, and Zhongfei Zhang.
\newblock Deep structured energy based models for anomaly detection.
\newblock In {\em ICML}, 2016.

\bibitem{zhang2022rethinking}
Junbo Zhang and Kaisheng Ma.
\newblock Rethinking the augmentation module in contrastive learning: Learning
  hierarchical augmentation invariance with expanded views.
\newblock In {\em CVPR}, 2022.

\bibitem{zhang2016colorful}
Richard Zhang, Phillip Isola, and Alexei~A Efros.
\newblock Colorful image colorization.
\newblock In {\em ECCV}, 2016.

\bibitem{zhou2017places}
Bolei Zhou, Agata Lapedriza, Aditya Khosla, Aude Oliva, and Antonio Torralba.
\newblock Places: A 10 million image database for scene recognition.
\newblock {\em IEEE TPAMI}, 40(6):1452--1464, 2017.

\bibitem{zong2018deep}
Bo Zong, Qi Song, Martin~Renqiang Min, Wei Cheng, Cristian Lumezanu, Daeki Cho,
  and Haifeng Chen.
\newblock Deep autoencoding gaussian mixture model for unsupervised anomaly
  detection.
\newblock In {\em ICLR}, 2018.

\end{thebibliography}
}

\appendix

\clearpage

\begin{center}{\bf {\Large Supplementary Material}}
\end{center}

In this supplementary material, we provide more experimental details and more experimental results (\emph{i.e.} per-class performance) on one-class CIFAR-100 (20 super-classes) and ImageNet-30, as well as more analysis on distribution-shifting/identity-preserving augmentations.

\section{Experimental Details}

\textbf{Hierarchical Augmentation.} We employ HA along the network to further prompt a higher concentration of inliers, in which deeper residual stages address stronger data augmentations. Following \cite{chen2020simple}, we use the combination of random resized crop, color jittering, gray-scale and horizontal flip with increasing augmentation strengths for $T_i~(i=1,2,3,4)$ to generate positive views. Table~\ref{tab:aug_res} shows the detailed augmentation configurations.

\begin{table}[htp]
\caption{Augmentation configurations for $T_i~(i=1,2,3,4)$. $RRC, CJ, GS, HF$ are short for random resized crop, color jittering, gray-scale and horizontal flip, respectively. $RRC(i, j)$ specifies the range of the cropped area and $CJ(b, c, s, h)$ specifies the range of brightness, contrast, saturation and hue.}
\label{tab:aug_res}
\resizebox{1.0\linewidth}{!}{
\begin{tabular}{c|c}
\toprule
$T_1$ & $RRC(0.75, 1), CJ(0.1, 0.1, 0.1, 0.025), GS, HF$ \\ \hline
$T_2$ & $RRC(0.54, 1), CJ(0.2, 0.2, 0.2, 0.050), GS, HF$ \\ \hline
$T_3$ & $RRC(0.30, 1), CJ(0.3, 0.3, 0.3, 0.075), GS, HF$ \\ \hline
$T_4$ & $RRC(0.08, 1), CJ(0.4, 0.4, 0.4, 0.100), GS, HF$ \\ 

\bottomrule
\end{tabular}
}
\end{table}

An extra projection head $g_i$ is additionally attached at the end of $res_i$ to down-sample and project the feature maps with the same shape as in the last stage $res_4$. Similar to \cite{zhang2022rethinking}, each $g_i$ consists of a series of down-sampling blocks and projection blocks. Table~\ref{tab:g_i} shows the detailed network structure. 

\begin{figure}[!t]
	\centering
	\includegraphics[width=0.48\textwidth]{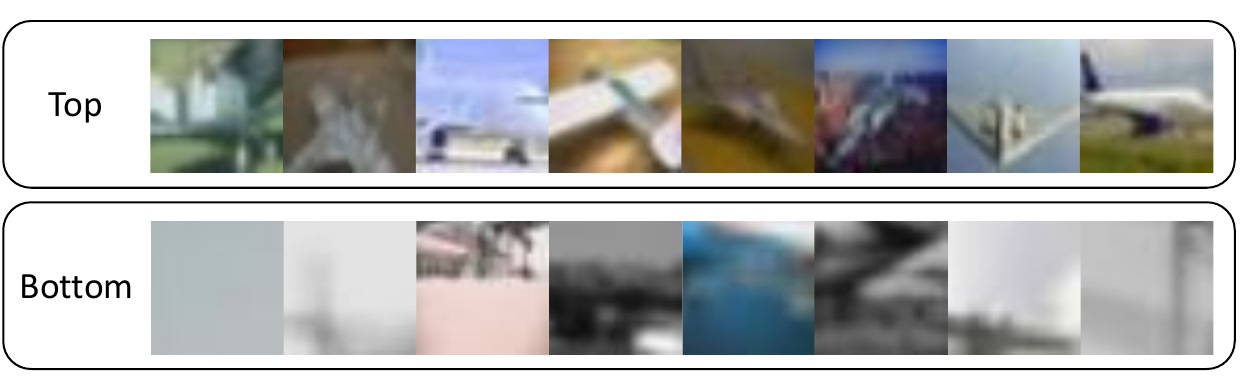}
	\caption{Illustration of augmented samples for the Plane class in CIFAR-10. Figures are from the same mini-batch during training and ranked according to the descent order of their $\omega_x$. Our soft mechanism enables us to identify the most likely inliers while suppress the potential outliers for a purified inlier concentration.}
	\label{fig:soft_weight}
\end{figure}

\textbf{Soft Aggregation.} In Fig.~\ref{fig:soft_weight}, we display two rows of the augmented views of inliers induced by standard data augmentation $\cal T$ as in CSI \cite{tack2020csi}. Notably, some views capture the main body of planes, whereas others are distracted by the background. It indicates that the generated views probably suffer from the semantic shift, and imposing such noisy inliers to be close reduces the purity of the inlier distribution.

\textbf{Outlier Exposure (OE) \cite{hendrycks2019oe}.} OE leverages an auxiliary dataset as outliers and enables anomaly detectors to generalize well to unseen anomalies. In this paper, we investigate the 80 Million Tiny Images dataset \cite{torralba200880} as the OE dataset with images from CIFAR-10 removed to make sure that the OE dataset and CIFAR-10 are disjoint. In practice, we use 300K random images\footnote{https://github.com/hendrycks/outlier-exposure} and observe that only a small fraction of this dataset is sufficiently effective for AD. Meanwhile, from Table~\ref{tab:oe-ratio}, we observe the increasing performance with more outliers exposed. Additionally, in the case of no OE applied, we vary $\lvert {\cal D}_{\rm out} \rvert$ by randomly keeping some inliers not being rotated. Table~\ref{tab:oe-ratio} shows that we can benefit more from a larger size of ${\cal D}_{\rm out}$.

\begin{table}[htp]
    \centering
    \caption{Ablation w.r.t. OE and $\lvert {\cal D}_{\rm out} \rvert$ ratios on CIFAR-10.}
    \label{tab:oe-ratio}
    \begin{tabular}{cccccc}
    \toprule
     & 0\% & 25\% & 50\% & 75\% & 100\% \\ \hline 
     $\lvert {\cal D}_{\rm out} \rvert$ & / & 91.3 & 92.6 & 94.5 & 95.4 \\
     OE & 95.4 &  95.9 & 96.2 & 96.6 &  \textbf{96.9} \\ 
     \bottomrule
    \end{tabular}
\end{table}

\begin{table*}
\caption{Per-class AUROC scores on one-class CIFAR-100 (20 super-classes). Numbers in the first column indicate the super-class IDs. * denotes the results directly adopted from \cite{tack2020csi} and bold numbers denote the best results.}
\label{tab:per_class_cifar100}
\resizebox{1.0\linewidth}{!}{
\begin{tabular}{ccccccccc}
 \toprule
 & OC-SVM* \cite{scholkopf1999support} & Geom \cite{golan2018deep} & Rot*\cite{hendrycks2019using_self} & Rot+Trans*\cite{hendrycks2019using_self} & GOAD*\cite{bergman2020classification} & DROC \cite{sohn2020learning} & CSI \cite{tack2020csi} & \textbf{UniCon-HA} (Ours) \\ 
 \hline

0 & 68.4 & 74.7 & 78.6 & 79.6 & 73.9 & 82.9 & 86.3  & \textbf{89.8} \\
1 & 63.6 & 68.5 & 73.4 & 73.3 & 69.2 & 84.3 & 84.8  & \textbf{90.2} \\
2 & 52.0 & 74.0 & 70.1 & 71.3 & 67.6 & 88.6 & 88.9  & \textbf{94.4} \\
3 & 64.7 & 81.0 & 68.6 & 73.9 & 71.8 & 86.4 & 85.7  & \textbf{89.5} \\
4 & 58.2 & 78.4 & 78.7 & 79.7 & 72.7 & 92.6 & 93.7  & \textbf{96.3} \\
5 & 54.9 & 59.1 & 69.7 & 72.6 & 67.0 & 84.5 & 81.9  & \textbf{87.6} \\
6 & 57.2 & 81.8 & 78.8 & 85.1 & 80.0 & 73.4 & 91.8  & \textbf{93.0} \\
7 & 62.9 & 65.0 & 62.5 & 66.8 & 59.1 & 84.2 & 83.9  & \textbf{87.8} \\
8 & 65.6 & 85.5 & 84.2 & 86.0 & 79.5 & 87.7 & 91.6  & \textbf{94.0} \\
9 & 74.1 & 90.6 & 86.3 & 87.3 & 83.7 & 94.1 & 95.0  & \textbf{97.1} \\
10& 84.1 & 87.6 & 87.1 & 88.6 & 84.0 & 85.2 & \textbf{94.0}  & 92.2 \\
11& 58.0 & 83.9 & 76.2 & 77.1 & 68.7 & 87.8 & 90.1  & \textbf{90.5} \\
12& 68.5 & 83.2 & 83.3 & 84.6 & 75.1 & 82.0 & 90.3  & \textbf{93.4} \\
13& 64.6 & 58.0 & 60.7 & 62.1 & 56.6 & 82.7 & 81.5  & \textbf{86.9} \\
14& 51.2 & 92.1 & 87.1 & 88.0 & 83.8 & 93.4 & 94.4  & \textbf{97.2} \\
15& 62.8 & 68.3 & 69.0 & 71.9 & 66.9 & 75.8 & \textbf{85.6}  & 84.2 \\
16& 66.6 & 73.5 & 71.7 & 75.6 & 67.5 & 80.3 & 83.0  & \textbf{90.8} \\
17& 73.7 & 93.8 & 92.2 & 93.5 & 91.6 & 97.5 & 97.5  & \textbf{98.1} \\
18& 52.8 & 90.7 & 90.4 & 91.5 & 88.0 & 94.4 & 95.9  & \textbf{98.0} \\
19& 58.4 & 85.0 & 86.5 & 88.1 & 82.6 & 92.4 & 95.2  & \textbf{96.7} \\ \hline
Mean& 63.1 & 78.7 & 77.7 & 79.8 & 74.5 & 86.5 & 89.6& \textbf{92.4} \\

\bottomrule

\end{tabular}
}
\end{table*}

\begin{table*}
\caption{Per-class AUROC scores on one-class ImageNet-30. Numbers in the first and fourth rows indicate the class IDs. Bold numbers denote the best results.}
\label{tab:per_class_imagenet30}
\resizebox{1.0\linewidth}{!}{
\begin{tabular}{cccccccccccccccc}
 \toprule

    & 0 & 1 & 2 & 3 & 4 & 5 & 6 & 7 & 8 & 9 & 10 & 11 & 12 & 13 & 14  \\ \hline 

    CSI \cite{tack2020csi} & 85.9 & \textbf{99.0} & \textbf{99.8} & 90.5 & 95.8 & 99.2 & 96.6 & 83.5 & 92.2 & 84.3 & \textbf{99.0} & 94.5 & 97.1 & 87.7 & 96.4 \\ 

    \textbf{UniCon-HA} & \textbf{87.3} & 98.7 & \textbf{99.8} & \textbf{93.1} & \textbf{96.4} & \textbf{99.3} & \textbf{97.5} & \textbf{88.4} & \textbf{94.3} & \textbf{89.2} & 98.9 & \textbf{95.3} & \textbf{97.4} & \textbf{90.0} & \textbf{96.7} \\

    \bottomrule
    
    & 15 & 16 & 17 & 18 & 19 & 20 & 21 & 22 & 23 & 24 & 25 & 26 & 27 & 28 & 29 \\ \hline

    CSI \cite{tack2020csi} & 84.7 & \textbf{99.7} & 75.6 & 95.2 & 73.8 & \textbf{94.7} & 95.2 & \textbf{99.2} & 98.5 & 82.5 & \textbf{89.7} & 82.1 & 97.2 & 82.1 & 97.6 \\
    \textbf{UniCon-HA} & \textbf{85.8} & 99.5 & \textbf{83.9} & \textbf{95.3} & \textbf{79.8} & 94.5 & \textbf{95.4} & 98.8 & \textbf{98.7} & \textbf{84.8} & 89.2 & \textbf{87.1} & \textbf{97.4} & \textbf{86.8} & \textbf{97.9} \\

    \bottomrule

\end{tabular}
}
\end{table*}

\begin{table}[htp]
\caption{The structure of the projection head $g_i$.}
\label{tab:g_i}
\resizebox{1.0\linewidth}{!}{
\begin{tabular}{c|c|c|c}
    \hline
    \#   &\multicolumn{2}{c|}{Down-sampling blocks}            & Projection blocks          \\  \hline
    \multirow{5}{*}{$g_1$}    & \multirow{2}{*}{SepConv}  &  Conv, Conv, BN, ReLU      &   \multirow{5}{*}{Linear, ReLU, Linear}    \\
        &  &  Conv, Conv, BN, ReLU                      &              \\   \cline{2-3}
    &  \multicolumn{2}{c|}{SepConv}              &           \\  
        &  \multicolumn{2}{c|}{SepConv}              &             \\  
        &  \multicolumn{2}{c|}{Conv, BN, ReLU, AvgPool}              &               \\  \hline
    \multirow{3}{*}{$g_2$}     &  \multicolumn{2}{c|}{SepConv}              &     \multirow{3}{*}{Linear, ReLU, Linear}            \\
    &  \multicolumn{2}{c|}{SepConv}              &                   \\
    &  \multicolumn{2}{c|}{Conv, BN, ReLU, AvgPool}              &              \\  \hline
    \multirow{2}{*}{$g_3$}     &  \multicolumn{2}{c|}{SepConv}              &         \multirow{2}{*}{Linear, ReLU, Linear}           \\
    &  \multicolumn{2}{c|}{Conv, BN, ReLU, AvgPool}              &            \\  \hline
    $g_4$     &  \multicolumn{2}{c|}{AvgPool}              &     Linear, ReLU, Linear           \\  \hline
\end{tabular}
}
\end{table}

\section{Per-class Results on One-class Settings}

Tables~\ref{tab:per_class_cifar100} and \ref{tab:per_class_imagenet30} present the AD results of our UniCon-HA on one-class CIFAR-100 (20 super-classes) and ImageNet-30, respectively. Clearly, our method outperforms the other state-of-the-art methods \cite{tack2020csi, golan2018deep, hendrycks2019using_self, bergman2020classification, sohn2020learning}, which also utilize transformations to create virtual outliers on most classes. 

Though sharing the same spirit of creating virtual outliers, we develop a completely different way of exploiting those outliers. Recall that a good representation distribution for AD
requires: (a) a compact distribution for inliers and (b) a dispersive distribution for (virtual) outliers. Both the requirements are only partially considered in the previous literature \cite{tack2020csi, golan2018deep, hendrycks2019using_self, bergman2020classification, sohn2020learning} with sub-optimal results obtained, while we explicitly encourage the concentration of inliers and the dispersion of outliers as our training objective. Interestingly, our method is free from any auxiliary branches to differentiate the specific types of transformations, outside of the commonly adopted transformation (\emph{e.g.} rotation) prediction based on a classifier for AD.

\section{Analysis on Augmentations}

Following CSI~\cite{tack2020csi}, we try to remove or convert-to-shift identity-preserving augmentations $\cal T$, including random resized crop, color jittering, horizontal flip and gray-scale. Table~\ref{tab:oc-cifar10-t-identity} confirms the observations from CSI: (1) treating $\cal T$ as distribution-shifting augmentations leads to a sub-optimal solution as these augmentations shift the original distribution less than rotation does, increasing false negative samples; (2) removing any augmentations from $\cal T$ degrades performance, showing the importance of identity-preserving augmentations to generating diverse positive views, where random crop is the most influential.

\begin{table}[htp]
    \centering
    \caption{Ablation study w.r.t. augmentations on CIFAR-10.}
    \label{tab:oc-cifar10-t-identity}

    \resizebox{0.95\linewidth}{!}{
    
    \begin{tabular}{ccccccc}
    \toprule
          & Base &  & Crop & Color & Flip & Gray \\  \midrule

      \multirow{2}{*}{CSI~\cite{tack2020csi}} & \multirow{2}{*}{94.3} & +shift  & 85.4 & 87.3 & 86.2 & 88.7  \\ \cline{3-7}  
        &  & -remove & 88.0 & 90.2 & 93.6 & 93.7 \\  \midrule

      \multirow{2}{*}{Ours} & \multirow{2}{*}{\textbf{95.4}} & +shift & 84.6 & 90.4 & 87.4 & 92.0   \\ \cline{3-7}  
          &  & -remove & 90.8 & 91.5 & 94.2 & 94.9 \\  
           
    \bottomrule
    \end{tabular}
    }

\end{table}

\end{document}